\documentclass{ifacconf}
\usepackage{subfig}

\usepackage{graphicx}      
\usepackage{natbib}        
\usepackage{booktabs}     
\usepackage{amsmath}      
\usepackage{amssymb}      

\newcommand{\f}{\mathbf{f}}

\newcommand{\g}{\mathbf{g}}

\newcommand{\p}{\mathbf{p}}
\newcommand{\q}{\mathbf{q}}

\newcommand{\R}{\mathbf{R}}
\renewcommand{\S}{\mathbf{S}}

\newcommand{\smallu}{\mathbf{u}}

\newcommand{\vel}{\mathbf{v}}

\newcommand{\x}{\mathbf{x}}

\newcommand{\z}{\mathbf{z}}

\newcommand{\btau}{\boldsymbol{\tau}}
\newcommand{\bomega}{\boldsymbol{\omega}}
\begin{document}
\begin{frontmatter}

\title{\LARGE \bf Non-Equilibrium MAV-Capture-MAV via Time-Optimal Planning and Reinforcement Learning\thanksref{footnoteinfo}} 

\thanks[footnoteinfo]{Sponsor and financial support acknowledgment
goes here. Paper titles should be written in uppercase and lowercase
letters, not all uppercase.}
\thanks[footnoteinfo1]{Cooresponding author}
\author[First,Second]{Canlun Zheng} 
\author[Second,Third]{Zhanyu Guo} 
\author[Second]{Zikang Yin}
\author[Second]{Chunyu Wang}
\author[Second]{Zhikun Wang\thanksref{footnoteinfo1}}
\author[Second]{Shiyu Zhao}

\address[First]{College of Computer Science and Technology, Zhejiang University, Hangzhou, China.}
\address[Second]{WINDY Lab, Department of Artificial Intelligence, Westlake University, Hangzhou, China. (e-mail: zhengcanlun, yinzikang, wangchunyu, wangzhikun, zhaoshiyu@westlake.edu.cn)}
\address[Third]{Department of Electrical Engineering, California Institute of Technology, Pasadena, USA.  (e-mail: zguo2@caltech.edu)}

\begin{abstract}    

This paper addresses intercepting highly maneuverable micro aerial vehicles (MAVs) using a compact platform with a custom launcher. We compare Time-Optimal Planning, which generates aggressive energy-efficient trajectories, with Reinforcement Learning, which offers superior robustness to dynamic uncertainties. Simulations and real-world experiments validate the RL-based approach, demonstrating reliable capture under unstable, high-speed conditions.
\end{abstract}


\end{frontmatter}

\section{Introduction}

This work focuses on the MAV-capture-MAV task, in which a pursuer MAV uses an onboard mechanism to neutralize or retrieve a target MAV\cite{vrba2022autonomous,chen2022anti,vrba2019onboard,mueller2013computationally}. 
From a research standpoint, this problem encompasses several challenging domains, including robust perception, state estimation, and aggressive control. 
From an application standpoint, the proliferation of commercial MAVs has introduced significant security and privacy risks due to potential misuse. Consequently, developing effective counter-MAV systems is imperative for public safety.

Previous work has addressed capture mechanisms, trajectory generation, and high-speed control. Our group has contributed vision-based MAV detection, bearing-only motion estimation, and guidance law design.

Previous studies have explored various capture mechanisms \cite{chen2022anti,liu2022safely}, trajectory generation algorithms \cite{wang2024impact}, and high-speed control schemes \cite{song2021autonomous}. 
Our group has contributed vision-based MAV detection\cite{zheng2022detection,zheng2024keypoint,guo2024global}, bearing-only motion estimation\cite{ning2024bearing,zheng2023optimal}, and guidance law design\cite{li2022three}.
However, a common limitation of existing approaches is their restriction to low-speed scenarios; for instance, our previous system was limited to target speeds of 4 m/s \cite{zheng2023optimal}. 
In contrast, modern high-performance MAVs routinely exceed these velocities. This speed gap stems largely from conventional control strategies, which often rely on simplified dynamics to ensure stable perception. Therefore, novel control methodologies are required to effectively engage highly maneuverable targets.

\begin{figure}
\centering
\includegraphics[width=1\linewidth]{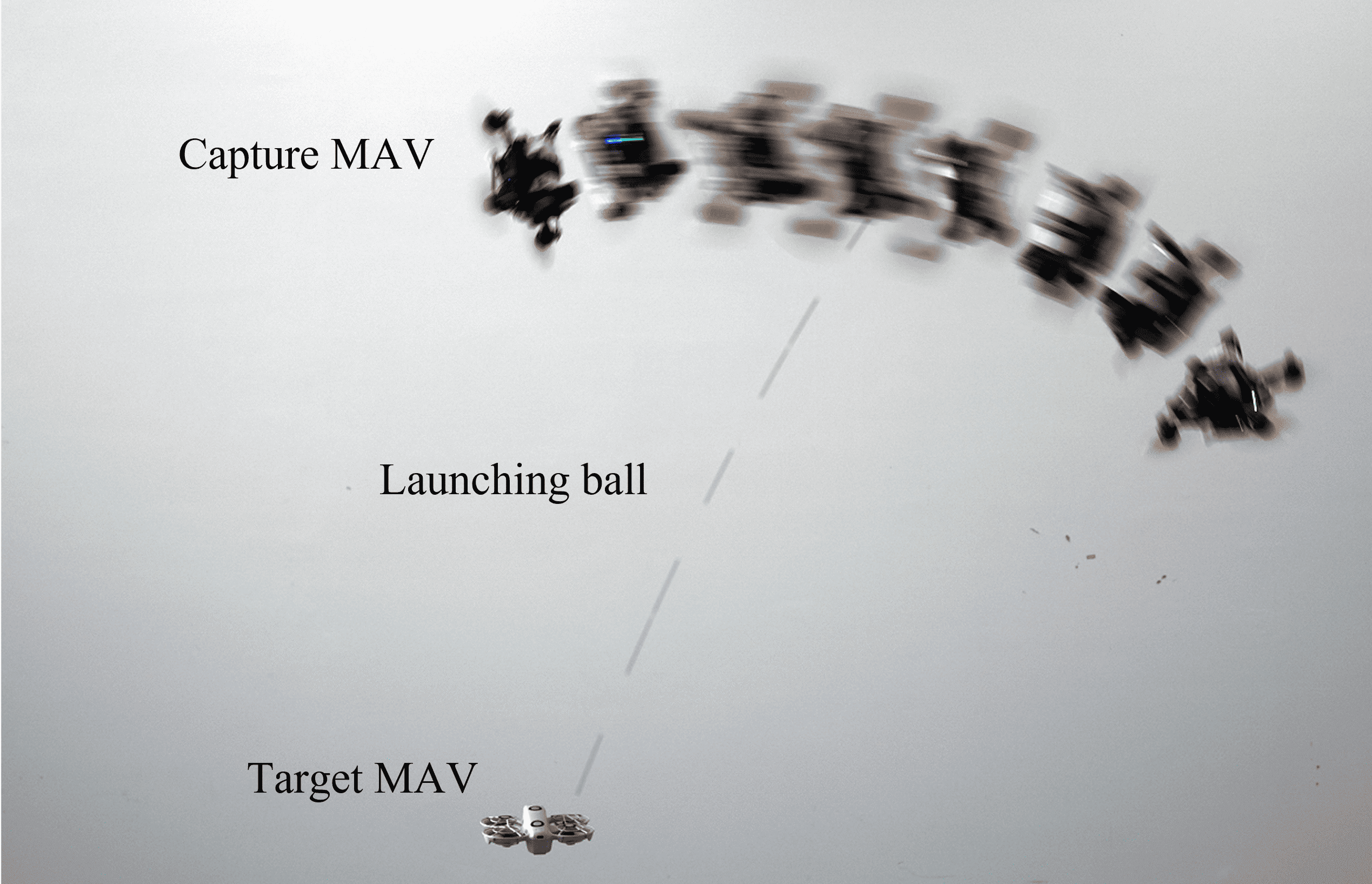} 
\caption{The capture MAV can launch a rubber ball in a non-equilibrium attitude to hit the target MAV.}\label{fig_ball_launching}
\end{figure}

Related research fields provide useful insights but differ fundamentally from our problem. Pursuit-evasion and guidance studies often assume that capture is complete once proximity is achieved\cite{jeon2016impact}, ignoring the physical process of launching an object toward the target.
Drone racing emphasizes time-optimal flight \cite{foehn2021time,kaufmann2023champion}, but usually with stationary gates or predefined waypoints, unlike unpredictable targets in capture. Ball-catching tasks \cite{su2017catching} exploit predictable trajectories, while MAV capture must cope with complex, evasive motion. Similarly, 
end-effector designs range from high-precision, low-agility manipulators\cite{kim2018stabilizing} to more maneuverable nets\cite{meng2018net}, yet both strongly constrain MAV dynamics. These comparisons highlight the uniqueness of our task, where a lightweight rubber-ball launcher (Fig.~\ref{fig_ball_launching}) allows active capture while preserving agility.

To tackle this challenge, we formulate MAV capture with a fixed launch device as a high-dimensional time optimization problem. Unlike racing or grasping tasks, this problem lacks explicit intermediate or terminal state references. Existing control strategies, such as PID or MPC, are effective only for slow targets \cite{zheng2023optimal}. In this work, we present the first comprehensive study of agile MAV capture control, comparing Time-Optimal Planning (TOP) and Reinforcement Learning (RL). Our contributions are threefold. First, TOP demonstrates the ability to generate highly maneuverable, minimum-time trajectories offline, but it incurs heavy computational loads and requires extremely accurate tracking. Second, RL enables real-time operation and exhibits improved stability after training, offering better adaptability to dynamic changes despite a slight reduction in trajectory optimality. Finally, we develop a compact, highly maneuverable MAV platform integrated with a custom ball-launching device, on which RL achieves successful high-speed interceptions in real-world experiments.

\section{Related Work}

In drone capture, end-effector design critically determines overall system performance. While manipulator-based approaches achieve high precision through multiple degrees of freedom\cite{kim2018stabilizing}, they severely compromise drone agility and are thus unsuitable for high-speed interception. 
Conversely, net-based systems offer a larger capture envelope and improved maneuverability \cite{chakravarthy2020collision,meng2018net}, but traditional methods rely on passive deployment triggered by proximity \cite{ning2024real}.
This dependency makes capture success highly sensitive to the drone's terminal attitude and relative velocity, offering limited fault tolerance in dynamic environments.

To address these limitations, we shift to an active capture mechanism. Specifically, we utilize a lightweight, fixed-angle rubber-ball launcher mounted on the pursuer MAV. This configuration decouples the capture action from the vehicle's base motion, allowing aggressive maneuvering without the aerodynamic penalties of nets or the inertial constraints of manipulators. 
Controlling such a system introduces unique challenges: the controller must both guide the drone and precisely time the launch to account for ballistic projectile dynamics.

The core difficulty lies in formulating this as a unified control problem. 
Unlike standard ball-catching tasks with predictable or quasi-linear trajectories \cite{bouffard2012learning,su2017catching}, our task involves a highly evasive target with unknown future states. Furthermore, unlike drone racing which follows predefined waypoints \cite{foehn2021time,romero2022model}, our problem lacks explicit intermediate or terminal state references. 
The agent must simultaneously optimize its trajectory and predict the target's motion to ensure successful projectile interception at a future time.

Time-optimal planning (TOP) provides a theoretical benchmark for aggressive maneuvers. Recent advances decompose the problem into manageable sub-tasks, using sampling-based methods such as RRT* for path generation and minimum-snap trajectories for smoothness \cite{burke2020generating}. These trajectories are typically tracked by high-rate controllers, such as Linear Model Predictive Control (LMPC) or geometric controllers on SE(3)\cite{han2021fast}.
However, solving these optimizations in real time remains computationally prohibitive for highly dynamic engagements.

RL learns nonlinear mappings from high-dimensional states to actions, outperforming human pilots in drone racing \cite{song2023reaching,song2021autonomous}. We formulate capture as an MDP with states (relative position/velocity, pursuer attitude/rates), actions (thrust, launch trigger), and a reward minimizing interception time while penalizing misses. RL thus delivers the real-time robustness traditional optimization lacks.

In summary, shifting from passive net capture to active projectile launching moves the primary challenge from mechanical design to high-level decision and control.
While TOP provides an ideal baseline for trajectory aggressiveness, its computational demands hinder real-time deployment. This work therefore explores RL to approximate optimal behavior, enabling robust, high-speed interception in practice.

\section{Problem Formulation}

\begin{figure}
    \centering
    \subfloat[]{\includegraphics[width=0.4\linewidth]{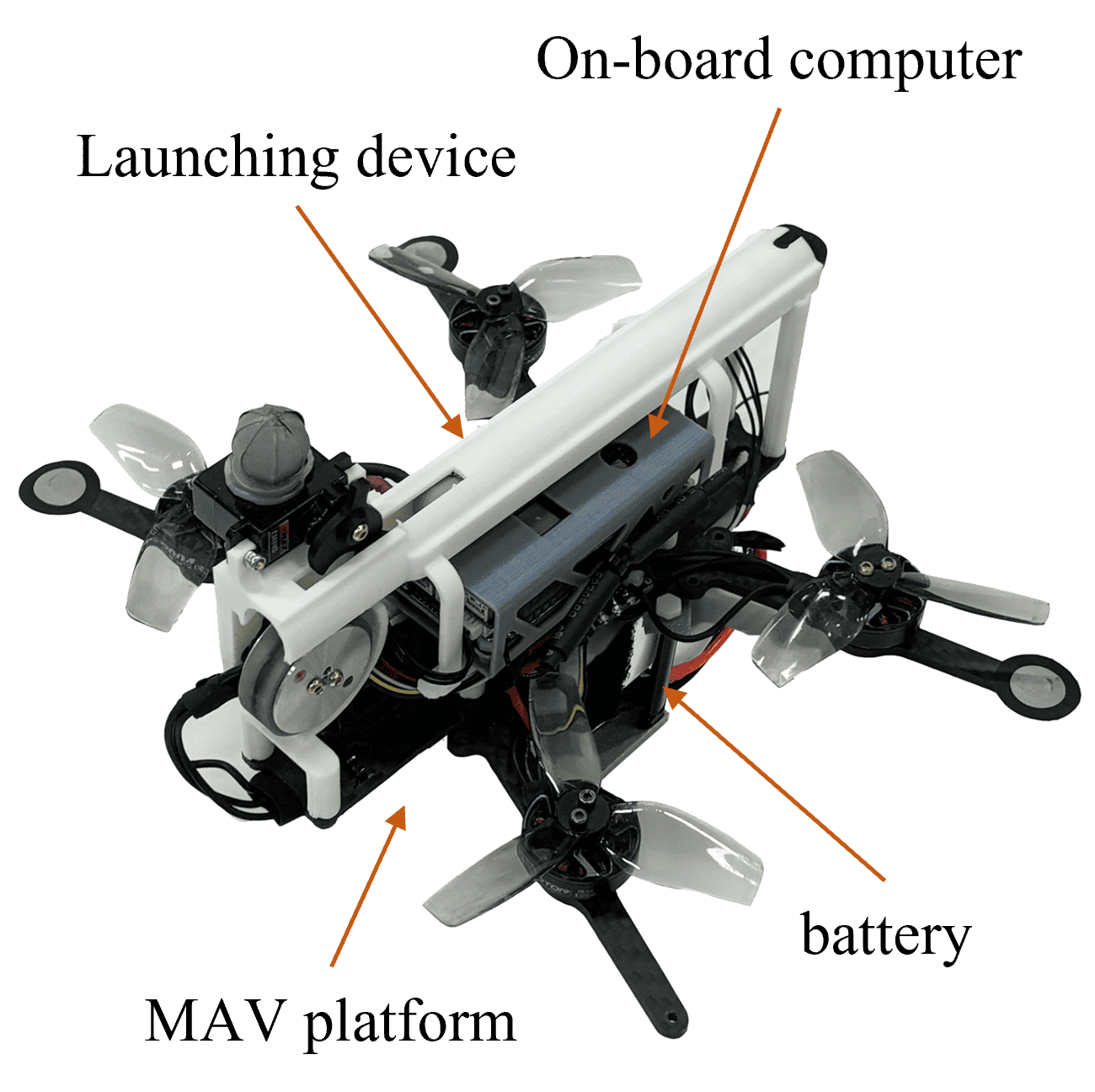}}
    \subfloat[]{\includegraphics[width=0.6\linewidth]{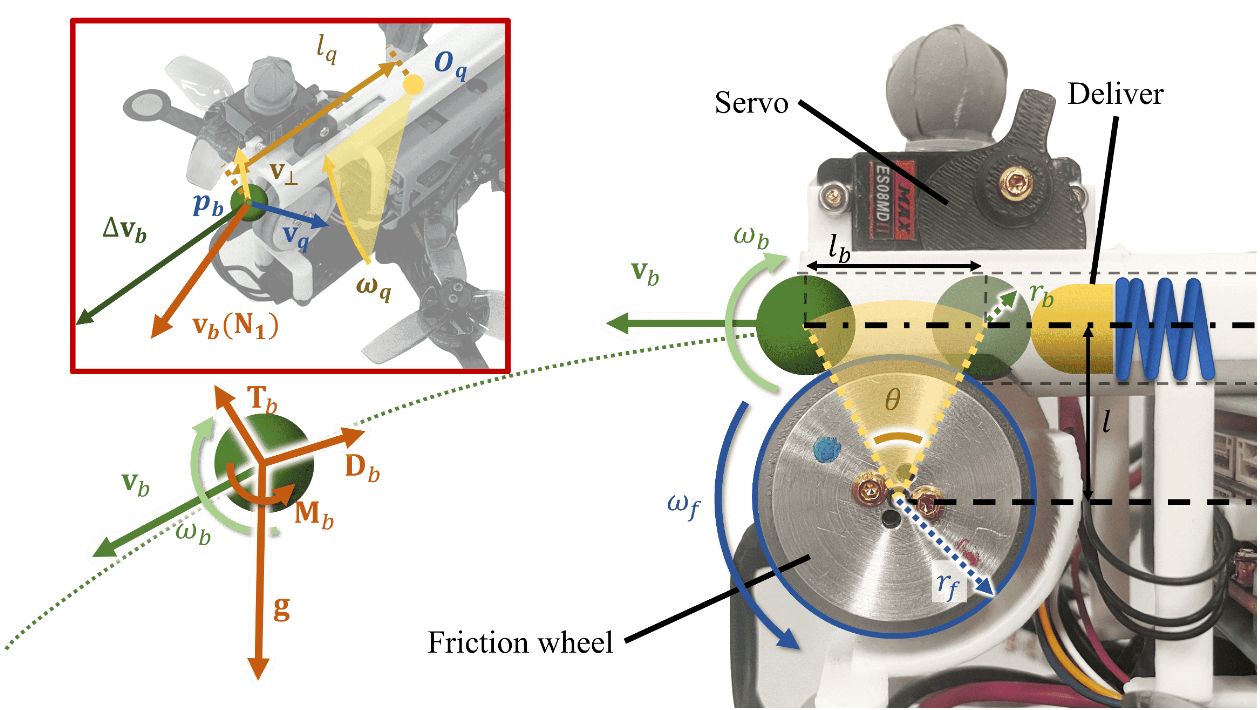}}
    \caption{The capture MAV platform. (a) shows the capture MAV. (b) shows the mechanical structure of the launch device, the dynamics of the flying ball, and the components of the launching velocity. 
    }\label{fig_shooting_device}
\end{figure}

Consider a capture MAV and a target MAV, as illustrated in Fig.~\ref{fig_ball_launching}. The capture MAV is equipped with a device to launch small rubber balls. Our goal is to design a control policy so that the capture MAV can rapidly approach the target MAV and launch a ball at the appropriate time. If the ball reaches a certain proximity to the target MAV, it is deemed successfully captured.
Additionally, as our first work on this topic, we focus solely on the control policy in this paper but do not consider perception or estimation, which will be addressed in the future.

\subsection{MAV Dynamics}

The state of the quadrotor comprises position $\p\in \mathbb{R}^3$, velocity $\vel\in \mathbb{R}^3$, unit quaternion $\q\in \mathbf{S}\mathbf{O}(3)$ representing body-to-world rotation, and body rates $\boldsymbol{\omega}\in \mathbb{R}^3$.
The dynamical equations are
\begin{align}\label{eqAVDynamics}
\begin{matrix}
\dot{\p} = \vel, & \dot{\vel} = \g + \frac{\mathbf{R}(\q)\e_3f_{\Sigma}}{m}, \\
\dot{\q} = \frac{\q}{2}\bigodot \begin{bmatrix}
    0 & \bomega^T
\end{bmatrix}^T, & \dot{\bomega} = \mathbf{J}^{-1}(\btau - \bomega\times \mathbf{J}\bomega),
\end{matrix}
\end{align}
where $\g$ is the gravity, $\e_3 = [0,0,1]^T$ is a unit vector, $\bigodot$ represents the Hamilton quaternion multiplication, $\mathbf{R}(\q)$ is the quaternion rotation matrix, $m$ and $\mathbf{J}$ are the MAV's mass and inertia, respectively. 
$f_{\Sigma}$ and $\btau$ are system inputs, which can be further decomposed as follows:
\begin{align}
f_{\Sigma}  = \sum f_i, \quad & 
\btau  = \begin{bmatrix}
	l/\sqrt{2}(f_1 + f_2 - f_3 -f_4)\\
	l/\sqrt{2}(- f_1 + f_2 + f_3 -f_4)\\
	c_{\tau}(f_1 - f_2 + f_3 -f_4)
\end{bmatrix},\label{eq_rotor}
\end{align}
where $f_i$ is the $i$th rotor's thrust, $l$ is the MAV's arm length, and $c_{\tau}$ is the rotor's torque constant.

The equations in \eqref{eqAVDynamics} can be written in a compact form as
$\dot{\x} = \f(\x,\smallu)$,
where $\x=[\p^T,\vel^T,\q^T,\bomega^T]^T\in \mathbb{R}^{13}$ is the state vector and $\smallu=[f_{\Sigma},\btau^T]^T\in\mathbb{R}^4$ is the input vector.

\subsection{Ball Dynamics}

The launch device comprises a spring deliverer, a servo trigger, and a friction wheel (as shown in Fig.~\ref{fig_shooting_device}(b)). The spring feeds balls to the servo, which releases them on command; the friction wheel then launches the ball via high-speed rotation. Let $l$ be the vertical distance from the friction wheel center to the launch tube center, \( r_f \) and \( r_b \) are the radius of the friction wheel and the ball, respectively, and the rotational speed of the friction wheel is \( \omega_f \). 
The ball's launch speed and angular velocity are
\begin{align}
v_{\rm launch} &= l_b/\delta t, \\
\omega_b &= (v_{\rm line} - v_{\rm launch})/r_b, \label{eq_launching_vel}
\end{align} 
where $l_b= 2\sqrt{(r_b + r_f)^2 - l^2}$ is the rolling distance. \(\delta t = (\theta + l_b/r_f)/\omega_f\) is the duration time, where \(\theta = 2 \mathrm{acos}(r_f+r_b,l)\) is the rolling angle of the friction wheel, \(v_{\rm line} = r_f\omega_f\) is the linear velocity of the ball.

After launching the ball,  the state of the ball is denoted as $\x_b =[\p_b^T,\vel_b^T,\omega_b]^T\in \mathbb{R}^7$, with the position $\p_b$, velocity $\vel_b$, and angular velocity $\omega_b$. The dynamics are
\begin{align}
\begin{matrix}
\dot{\p}_b = \vel_b, &  \dot{\vel}_b = \g + \frac{\mathbf{T}_b +\mathbf{D}_b}{m_b}, & \dot{\omega}_b = \frac{M_b}{J_b},  
\end{matrix}\label{eq_ball_dynamic}
\end{align}
where $m_b$ and $J_b$ are the ball's mass and inertia.

The lift force $\mathbf{T}_b\in \mathbb{R}^3$, drag force $\mathbf{D}_b\in \mathbb{R}^3$, and moment $M_b\in \mathbb{R}$ are defined as
\begin{align}
    \mathbf{T}_b  = {{\rm C}{\rm C}_l}\omega_b \e_b, ~  \mathbf{D}_b  = - \frac{ \rm{C}{\rm C}_d \vel_b}{\|\vel_b\|}, ~ M_b  = {{\rm C}{\rm C}_M 2 r_b} \omega_b,
\end{align}
where ${\rm C} = \frac{1}{2}{\rm \rho}\|\vel_b\|^2S$, $\rm{C}_l, \rm{C}_d$ and ${\rm C}_M$ are the aerodynamic coefficients, ${S}=\pi r_b^2$ is the cross-sectional area of the ball.
The dynamics can be described in a compact form as 
$\dot{\x}_b = \f_b(\x_b)$.

\begin{figure}
    \centering
    \includegraphics[width=1\linewidth]{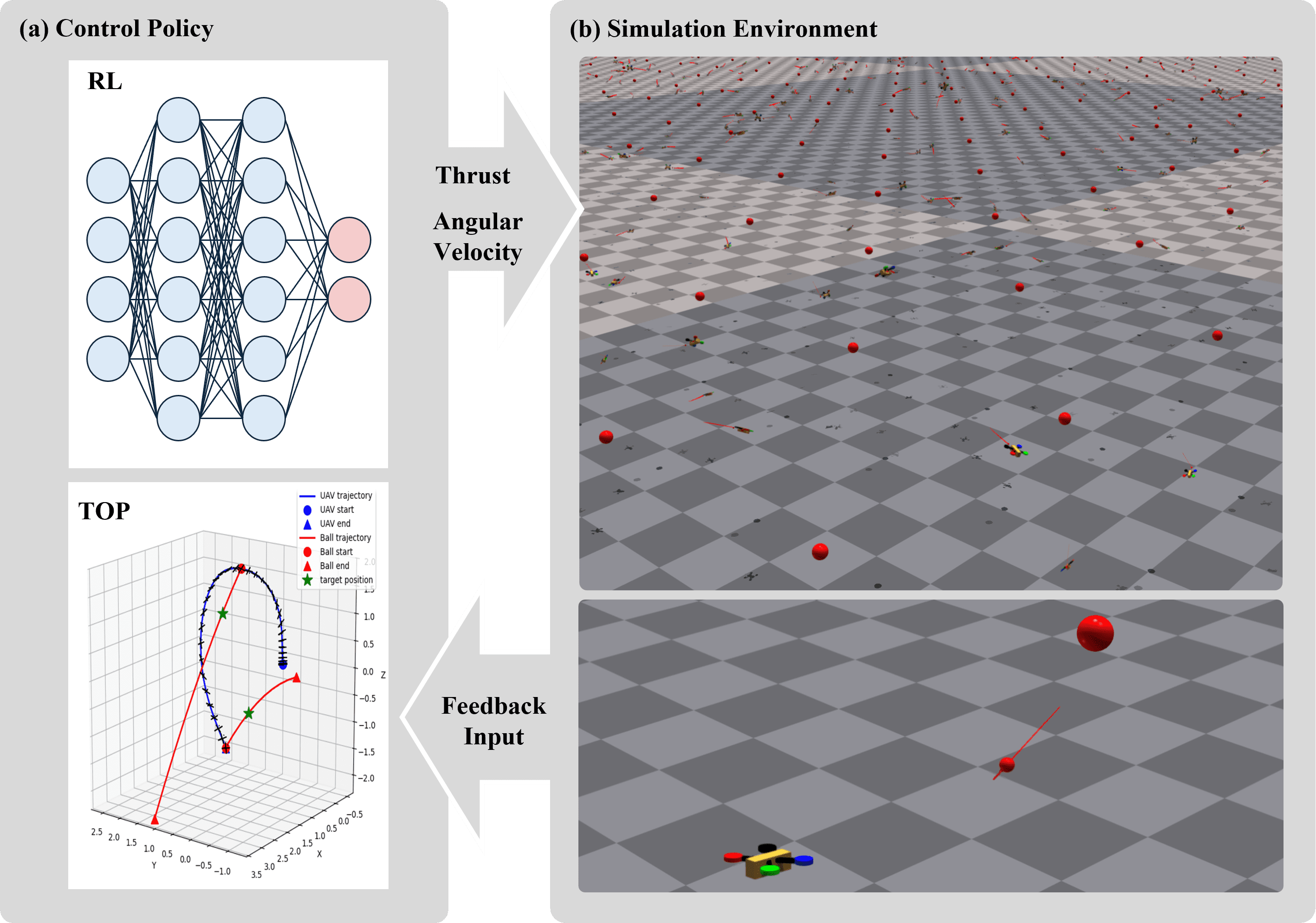}
    \caption{(a) is the control policy, including RL and TOP. (b) is the training simulation environment developed in Isaac Gym.}
    \label{fig_sys}
\end{figure}

\section{Time-Optimal Planning Method}

The objective is to minimize the total capture time $t_c = t_1 + t_2$, where $t_1$ is the flight time of the capture MAV from its initial state to the launch state, and $t_2$
is the flight time of the ball to the target. The optimization problem is formulated as
\begin{align}
&\min_{\{\smallu(k)\}_{k=0}^{N_1}, t_1,t_2} t_c\nonumber\\
\x(k+1) & = \x(k) + \frac{t_1}{N_1}\f(\x(k),\smallu(k)), \quad 0\leq k \leq {N_1} \label{eq_dicreuad}\\
\x_b(0) & = [\p_b({N_1})^T, \vel_b({N_1})^T,\omega_b({N_1})]^T,\label{eq_ball_init}\\
\x_b(k+1) & = \x_b(k) + \frac{t_2}{ N_2} \f_b\left(\x(k)\right) , \quad  0\leq k \leq {N_2} \label{eq_ball_dynamic_f},\\
d({N_2}) & = \|\p_b({N_2}) - \p_T({N_2})\|_2\leq \sigma_{d},\label{eq_hit_threhold},\\
\|\smallu(k)\|_\infty & \leq u_{\max}, \quad 0\leq k \leq {N_1}, \label{eq_control_constraint}\\
\p_z(k) & \geq z_{\min}, \quad 0\leq k \leq {N_1}, \label{eq_altitude_constraint}
\end{align}
Equation \eqref{eq_dicreuad} discretizes the MAV dynamics over $N_1$ steps, which is selected as 20 in our experiments; \eqref{eq_ball_init} sets the ball's initial state at launch; \eqref{eq_ball_dynamic_f} propagates the ball dynamics over $N_2$ steps; \eqref{eq_hit_threhold} enforces that the final ball–target distance is within threshold $\sigma_d$; and \eqref{eq_control_constraint}–\eqref{eq_altitude_constraint} impose input bounds and a minimum altitude $\z_{min}$. The launch state is defined as $\x({N_1})$.
The NLP is solved via IPOPT in CasADi, warm-started with the previous solution shifted by one step to exploit temporal continuity and accelerate convergence.


Second, the initial state of the ball is calculated in the following way
\begin{align}
    \x_b(0)=[\p_b({N_1})^T, \vel_b({N_1})^T,\omega_b({N_1})]^T,
\end{align}
 where $\p_b({N_1})$ and $\vel_b({N_1})$ are given by
\begin{align}
    \p_b(N_1) & = \p(N_1) + l\mathbf{R}\e_1, \nonumber\\
    \vel_b(N_1) & = \vel(N_1) + \Delta \vel_b + \vel_\bot,\label{eq_ball_vel_0}
\end{align}
where \(\Delta \vel_b= \R \e_1 v_{\mathrm{launch}} \); \(\vel_\bot = \R(t_1) \bomega(t_1)\times \e_1 l_g\) is the throwing velocity caused by the rotation of the capture MAV, and \( l \) is the offset distance from the launch point to the capture MAV's center of mass.

\section{Reinforcement Learning Method}

\subsection{Problem Setup}

To solve the MAV capture task under the RL framework, we model the task as a Markov Decision Process (MDP).
In MDP, the state is $[\p^T, \vel^T,{\rm vec}(\R_q)^T, \omega^T, \p_T^T, \vel_T^T]^T \in \mathbb{R}^{24}$, and the action is \([f_\Sigma, \boldsymbol{\omega}^T]^T \in \mathbb{R}^{4}\). 
The simulation dynamics follow \eqref{eqAVDynamics} and \eqref{eq_ball_dynamic}. 

The policy is a fully connected network with two hidden layers of 512 nodes each, mapping states to actions.
The goal is to find an optimal policy that can maximize the following average reward objective function \cite[Chapter~9]{zhao2024RLBook}:
\begin{align}
\pi^* = \arg\max_{\pi} \mathbb{E}\left[\sum_{k=0}^{\infty} \gamma^k r(k)\right].
\end{align}
We optimize the policy using PPO \cite{schulman2017proximal} to maximize the discounted return with discount factor \( \gamma = 0.99 \). The reward function \( r \) is detailed below.



\subsection{Reward Function}

Since directly optimizing the total capture time yields sparse rewards, we adopt a distance-based dense reward to approximate time minimization. The reward comprises four terms.

\textbf{1) Pursuit distance reward:} The distance \( d = \|\p - \p_T\| \) should lie within \( [d_{\min}, d_{\max}] \), where \( d_{\max} \) ensures proximity for launching and \( d_{\min} \) prevents collision. We define
\begin{align}
r_d = \frac{1}{1 + \max(\min(d, d_{\max}), d_{\min})},
\end{align}
with \( d_{\min}=1\,\text{m} \) and \( d_{\max}=1.5\,\text{m} \).

\textbf{2) Ball interception reward:} The minimum ball–target distance over the predicted trajectory within one second after launch is \( d_b = \min_{\tau\in[0,1]} \|\p_b(\tau) - \p_T(\tau)\| \), computed via \eqref{eq_ball_dynamic} and \eqref{eq_ball_vel_0}. The reward is
\begin{align}
r_b = \frac{1}{1 + d_b}.
\end{align}

\textbf{3) Angular velocity penalty:} Excessive spin may cause missed launch opportunities. We penalize body rates via
\begin{align}
r_{\bomega} = \frac{1}{1 + \|\bomega\|^2}.
\end{align}

\textbf{4) Altitude safety reward:} To prevent ground collision, we define
\begin{align}
r_a = \begin{cases}
2p_z, & p_z < 0.5,\\
1,    & p_z \geq 0.5.
\end{cases}
\end{align}

\textbf{Total reward:} The composite reward is
\begin{align}
r = r_a r_d (r_b + r_{\bomega}) + r_d,
\end{align}
where the multiplicative structure encourages joint optimization of all terms, and the additive \( r_d \) provides a baseline distance incentive.

\section{Experiments}\label{sec_experiments}

\begin{figure}
\centering
\includegraphics[width=1\linewidth]{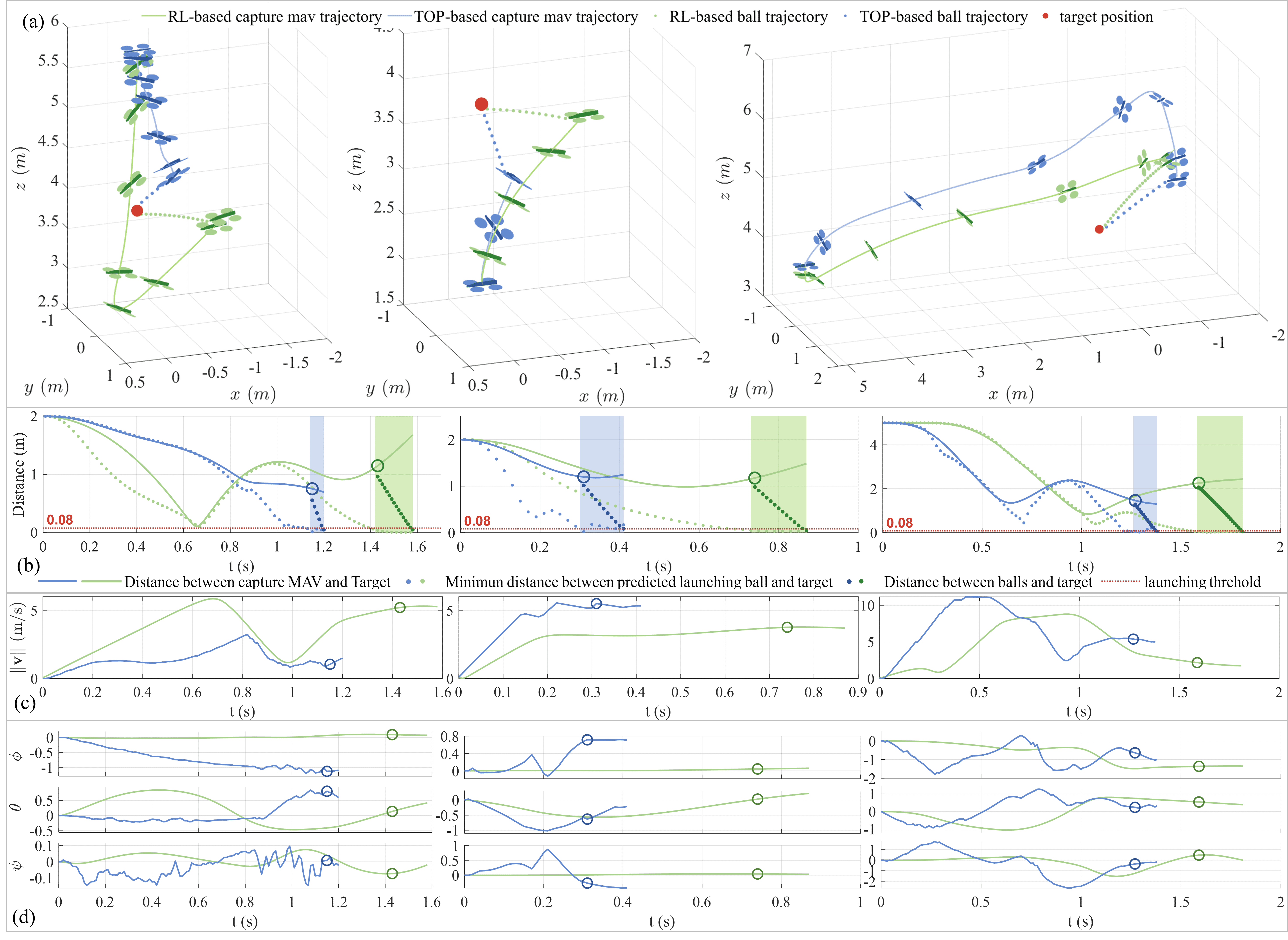}
\caption{Scenario 1 simulation results. (a) Launching at a static target from three initial positions (above, below, behind). (b) Distances: MAV–target, predicted ball–target minimum, and launched ball–target for both methods. (c) Capture MAV speeds. (d) Capture MAV attitudes. In (b)–(d), \(\circ\) denotes the launch instant.}\label{fig_sim1}
\end{figure}

\begin{figure}
\centering
\includegraphics[width=1\linewidth]{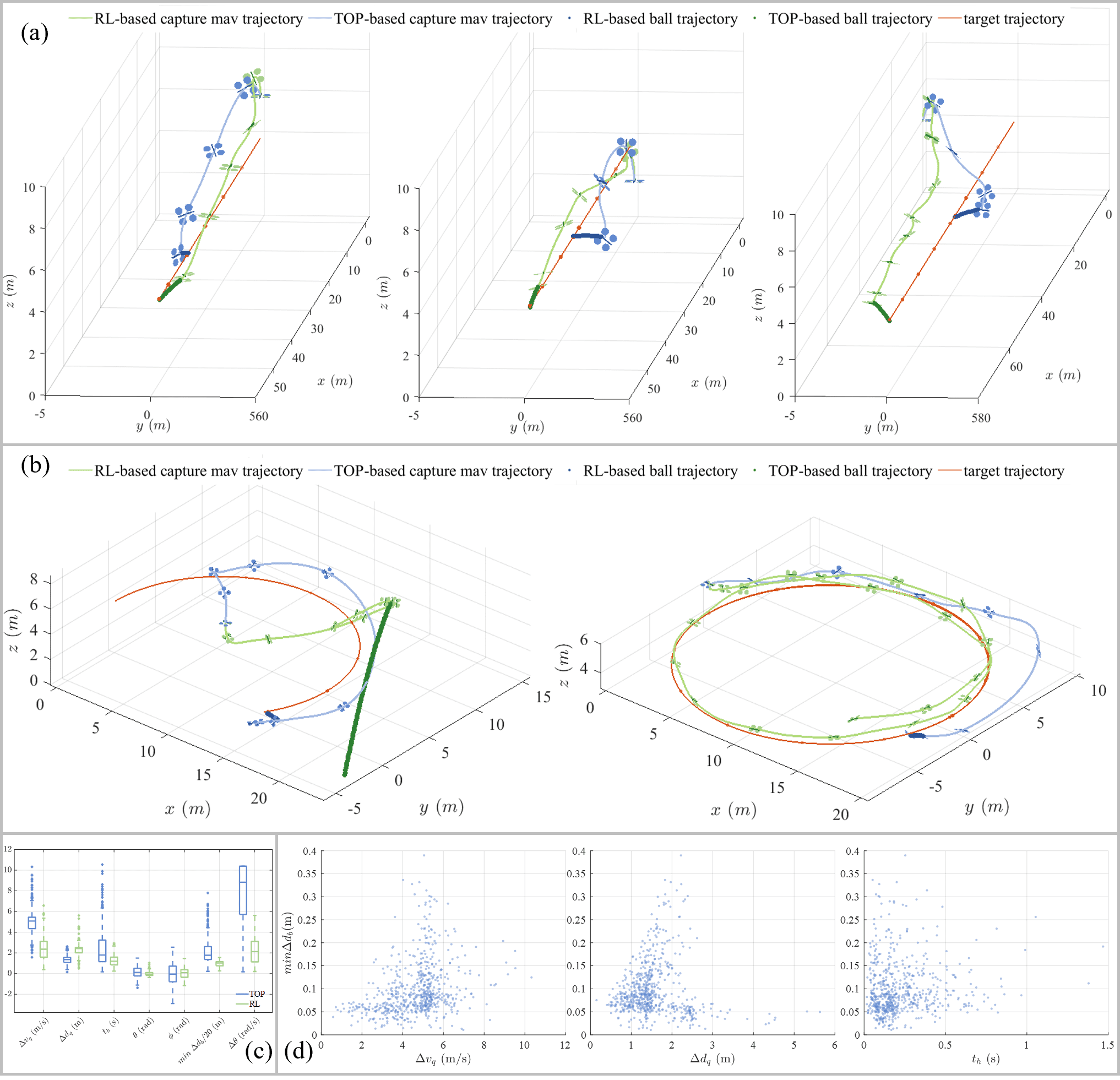}
\caption{  Scenario 2 simulation results. (a) Launching at a target moving at 2 m/s from four initial positions. (b) Launching at a target moving at 4 m/s in a circular trajectory from four initial positions. (c) Launch state statistics: relative speed, relative distance, flight time, \(\theta\), \(\phi\), miss distance, and angular velocity. (d) Miss distance versus relative speed, relative distance, and ball flight time.}\label{fig_sim2}
\end{figure}

We conducted simulations in Isaac Gym (Fig.~\ref{fig_sys}(b)) and real-world experiments. Two simulation scenarios were considered: a stationary target and a target moving at constant or time-varying speeds. The real-world experiment tested RL-based capture of a static target.

Key RL hyperparameters are listed in Table~\ref{tab:ppo_hyperparameters}. In real-world tests, the MAV state is provided by a Vicon motion capture system at 100 Hz. The trained policy runs onboard at 100 Hz, enabling fully autonomous flight without external computation.
The trained neural network policy is executed onboard the MAV, where control inference is performed at 100~Hz. 
All computations for policy inference are carried out on the onboard computer, allowing the system to operate fully autonomously without reliance on external computation during flight.

\begin{table}
\centering
\caption{Hyperparameters for PPO Training}
\label{tab:ppo_hyperparameters}
\begin{tabular}{ll}
\toprule
\textbf{Parameter} & \textbf{Value} \\
\midrule
Clipping parameter & 0.2 \\
Target KL divergence & 0.03 \\
Discount factor & 0.99 \\
GAE parameter & 0.95 \\
Learning rate & $3 \times 10^{-4}$ \\
Episode length & 1000 \\
Training epochs & 300 \\
Number of parallel environments & 2048 \\
\bottomrule
\end{tabular}
\end{table}

\subsection{Scenario 1: Stationary Target MAV}

The target MAV hovers at \( \p_{T} = [0,0,4] \). 
The capture MAV starts from three relative positions: top (\(\p = [0,0,6]\)), bottom (\(\p = [0,0,2]\)), and front (\(\p = [-5,0,4]\)). 
Capture is declared when the ball–target distance falls below \( \sigma_d = 0.1\,\text{m} \).

As shown in Fig.~\ref{fig_sim1}, both TOP and RL successfully capture the target but exhibit distinct behaviors. 
TOP achieves shorter flight times with more aggressive attitude maneuvers, enabling rapid arrival at the launch state. However, this shortens the effective launch window, risking missed opportunities and requiring replanning. 
In contrast, RL produces smoother, lower-speed trajectories with a wider launch window. 
RL also shows a consistent preference for high-altitude attack, approaching from above and launching when conditions are favorable. This behavior improves aiming precision and shooting efficiency, mirroring the strategic advantage of elevated attack positions observed in natural predation.


\subsection{Scenario 2: Moving Target MAV}

We considered two sub-scenarios. In the first, the target MAV moves with a constant velocity \(\vel_{T} = [20,0,0]^T~{\rm m/s}\). In the second, it follows a circular trajectory with a speed of \(10~{\rm m/s}\) and an acceleration of \(10~{\rm m/s}^2\).

Fig.~\ref{fig_sim2}(a) shows the results for the constant-velocity case. The capture MAV starts from three positions relative to the target: top, bottom, and side. As in the stationary scenario, TOP produces more aggressive maneuvers and reaches the launch state faster than RL.

Fig.~\ref{fig_sim2}(b) shows the results for the circular-motion case, where the capture MAV starts from $\p$ = [10,0,6] and $\p$ = [0,0,6]. From the first position, RL fails because it launches from a longer distance, leading to significant target prediction error. TOP succeeds by launching closer to the target.

Fig.~\ref{fig_sim2}(c) summarizes statistics from 300 simulation trials. TOP achieves more aggressive launch attitudes and shorter capture times, whereas RL generates smoother trajectories and smaller miss distances. This suggests that flight stability contributes to higher capture success rates.

Fig.~\ref{fig_sim2}(d) shows that miss distance increases rapidly with relative speed, relative distance, and ball flight time. Higher relative speed narrows the launch window, while larger launch distances amplify target prediction uncertainty, both leading to greater miss distances.

\subsection{Real-world Experiments}

\begin{figure}
    \centering
    \includegraphics[width=1\linewidth]{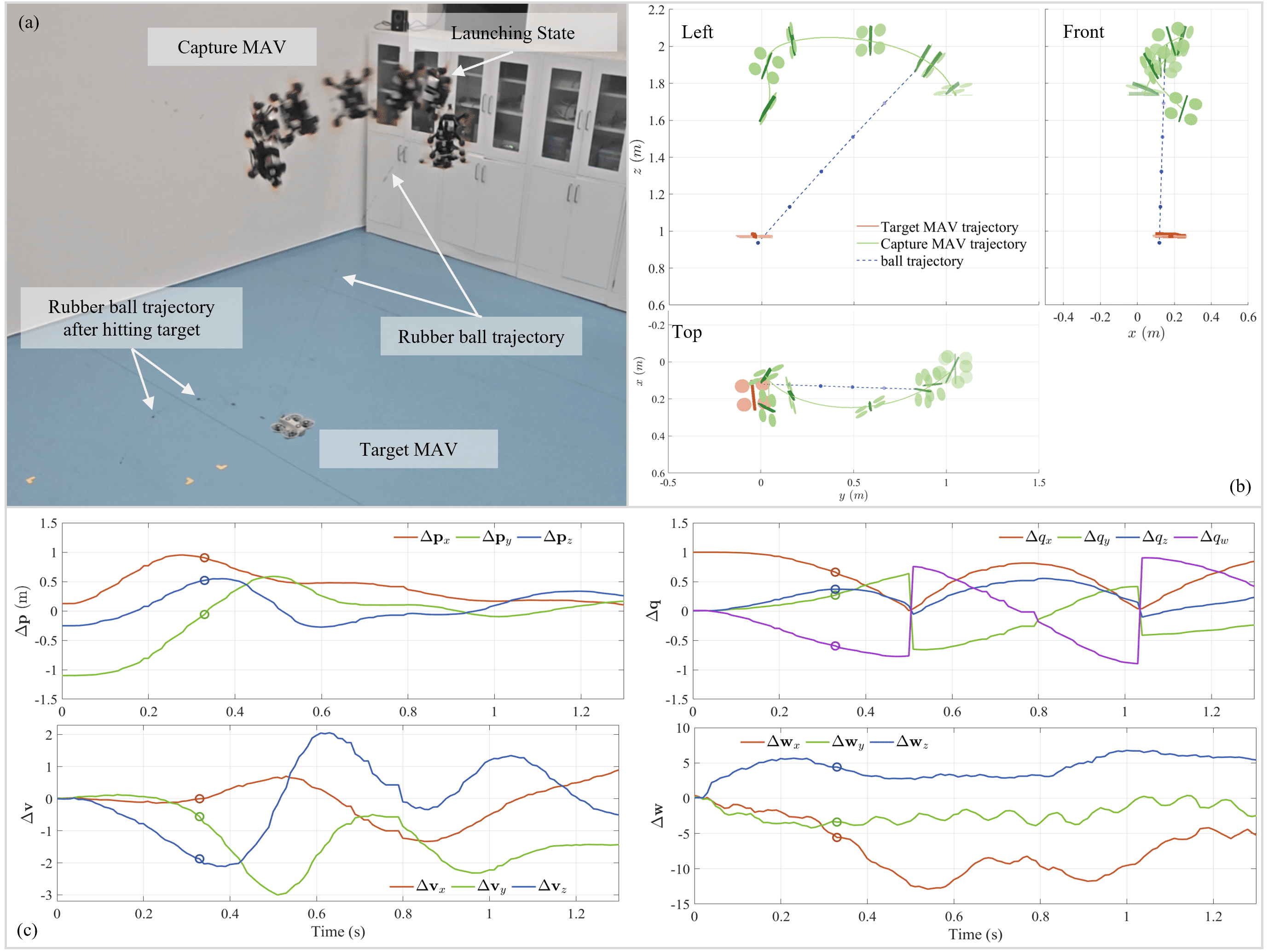}
    \caption{The real-world flight experimental results of RL-based MAV-capture-MAV.}
    \label{fig_exp}
\end{figure}

We conducted an indoor experiment with one capture MAV (as shown in Fig.~\ref{fig_shooting_device}(a)) and one target MAV (DJI nio). Only the RL method is tested since the high-demand computation power of TOP cannot be executed in real onboard systems. The experiment's results are shown in Fig.~\ref{fig_exp}. 
Fig.~\ref{fig_exp}(a) shows the capturing process during 1.3~second. Fig.~\ref{fig_exp}(b) shows the three views of the spatial relationship between the capture MAV and the target.
Fig.~\ref{fig_exp}(c) shows the relative position, quaternion, linear velocity, and angular velocity between the capture MAV and the target. 
The experiment also shows that the RL method prefers a high-altitude attack strategy, as shown in simulation experiments.

\section{Conclusion}

In this work, we presented a comparative study of TOP and RL for intercepting agile MAVs using a compact platform with a lightweight launcher. While TOP generates highly aggressive trajectories, its computational intensity limits real-time applicability. Conversely, RL demonstrates superior robustness and real-time capability, successfully transferring from simulation to physical hardware. This study validates the effectiveness of an active capture mechanism and establishes RL as a viable approach for high-speed interception tasks. Future work will extend this framework to multi-agent cooperation and address perceptual uncertainties.


\bibliography{main}             

\end{document}